\documentclass[11pt,a4paper]{article}
\usepackage[hyperref]{acl2019}
\usepackage{times}
\usepackage{latexsym}

\usepackage{url}

\aclfinalcopy 
\def\aclpaperid{1895} 

\usepackage{amsmath,amsfonts,bm}

\def\eqref#1{equation~\ref{#1}}

\def\1{\bm{1}}

\def\vw{{\bm{w}}}

\DeclareMathAlphabet{\mathsfit}{\encodingdefault}{\sfdefault}{m}{sl}
\SetMathAlphabet{\mathsfit}{bold}{\encodingdefault}{\sfdefault}{bx}{n}

\DeclareMathOperator*{\argmax}{arg\,max}
\DeclareMathOperator*{\argmin}{arg\,min}

\usepackage{url}
\usepackage{graphicx}
\usepackage{booktabs}
\usepackage{multirow}
\usepackage{mathtools} 
\usepackage{paralist}
\usepackage{csquotes} 
\usepackage{makecell}
\usepackage[export]{adjustbox}
\usepackage[bottom]{footmisc}
\usepackage{caption}
\usepackage{subcaption}

\title{CoDraw: Collaborative Drawing as a Testbed for\\ Grounded Goal-driven Communication}

\newcommand*\samethanks[1][\value{footnote}]{\footnotemark[#1]}
\author{Jin-Hwa Kim\thanks{\hspace{0.6em}The first two authors contributed equally to this work.} \\
SK T-Brain\thanks{\hspace{0.6em}Work performed while the authors were interns at Facebook AI Research.} \\
\texttt{jnhwkim@sktbrain.com} \\
\And
Nikita Kitaev\samethanks[1] \\
University of California, Berkeley\samethanks[2] \\
\texttt{kitaev@cs.berkeley.edu} \\
\AND
Xinlei Chen, Marcus Rohrbach \\
Facebook AI Research \\
\texttt{\{xinleic,mrf\}@fb.com} \\
\And
Byoung-Tak Zhang \\
Seoul National University \\
\texttt{btzhang@bi.snu.ac.kr} \\
\AND
Yuandong Tian \\
Facebook AI Research \\
\texttt{yuandong@fb.com} \\
\And
Dhruv Batra \& Devi Parikh \\
Georgia Institute of Technology, Facebook AI Research \\
\texttt{\{parikh,dbatra\}@gatech.edu}
}

\newenvironment{packed_itemize}{
\begin{list}{\labelitemi}{\leftmargin=2em}
\vspace{-6pt}
 \setlength{\itemsep}{0pt}
 \setlength{\parskip}{0pt}
 \setlength{\parsep}{0pt}
}{\end{list}}

\usepackage{xspace}

\newcommand{\Teller}[0]{\text{Teller}\xspace}
\newcommand{\Drawer}[0]{\text{Drawer}\xspace}
\newcommand{\CoDraw}[0]{\text{CoDraw}\xspace}

\newcommand{\myparagraph}[1]{\noindent\textbf{#1.}}

\makeatletter
\DeclareRobustCommand\onedot{\futurelet\@let@token\@onedot}
\def\@onedot{\ifx\@let@token.\else.\null\fi\xspace}

\def\eg{\emph{e.g}\onedot}

\makeatother

\date{}
\begin{document}

\maketitle

\begin{abstract}
In this work, we propose a goal-driven collaborative task that combines language, perception, and action.
Specifically, we develop a \emph{Co}llaborative image-\emph{Draw}ing game between two agents, called \emph{\CoDraw}.
Our game is grounded in a virtual world that contains movable clip art objects.
The game involves two players: a \emph{Teller} and a \emph{Drawer}.
The \Teller sees an abstract scene containing multiple clip art pieces in a 
semantically meaningful configuration, while the \Drawer tries to reconstruct the 
scene on an empty canvas using available clip art pieces. The two players communicate 
with each other using natural language. 
We collect the {\CoDraw} dataset of $\sim$10K dialogs consisting of $\sim$138K messages exchanged between human players. We define protocols and metrics to evaluate learned agents in this testbed, highlighting the need for a novel \emph{crosstalk} evaluation condition which pairs agents trained independently on disjoint subsets of the training data.
We present models for our task and benchmark them using both fully automated evaluation and by having them play the game live with humans.

\end{abstract}

\begin{figure}[ht!]
  \centering
  \includegraphics[width=\linewidth]{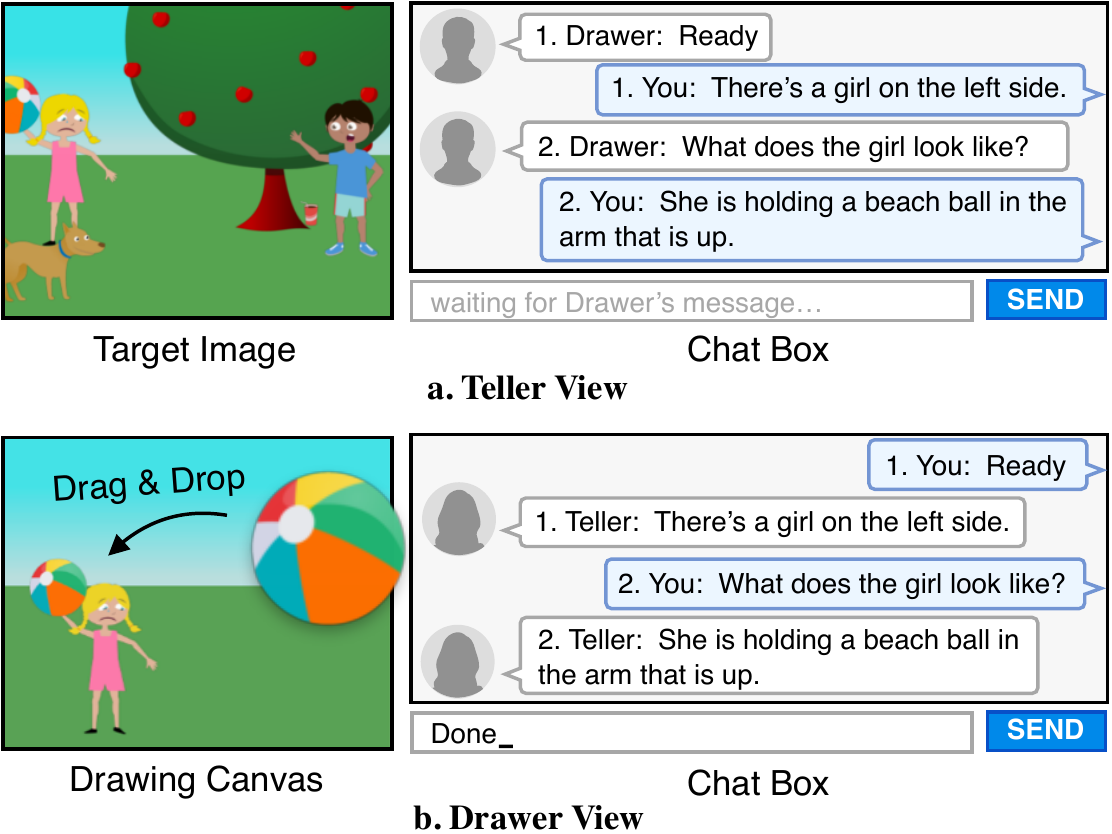}
  \caption{Overview of the proposed Collaborative Drawing (\CoDraw) task. The game consists of two players -- \Teller and \Drawer. The \Teller sees an abstract scene, while the \Drawer sees an initially empty canvas. Both players need to collaborate and communicate so that the \Drawer can drag and drop clip art objects to reconstruct the target scene that is only visible to the \Teller.
  }
  \label{fig:schema}
  \vspace{-0.7em}
\end{figure}

\section{Introduction}

Building agents that can interact with humans in natural language while perceiving and taking actions in their environments is one of the fundamental goals in artificial intelligence. 
To this end, it will be necessary to \emph{ground} language into perception and action~\citep{harnad1990symbol,barsalou1999perceptions},
where, \textit{e.g.}, nouns are connected to percepts and verbs relate to actions taken in an environment.
Some approaches judge machine understanding of language purely based on the ability to mimic particular human utterances, but this has limitations: there are many ways to express roughly the same meaning, and conveying the correct information is often more important than the particular choice of words.
An alternative approach, which has recently gained increased prominence, is to train and evaluate language capabilities in an \emph{interactive} setting, where the focus is on \emph{successfully communicating information} that an agent must share in order to achieve its goals.

\pagebreak 
In this paper, we propose the Collaborative Drawing (\CoDraw) task, which combines grounded language understanding and learning effective goal-driven communication into a single, unified testbed. This
task involves perception, communication, and actions in a
partially observable environment.
As shown in Figure~\ref{fig:schema}, our game is grounded in a virtual world constructed from clip art objects~\citep{Zitnick2013,Zitnick2013a}. Two players, \Teller and \Drawer, play the game.  The \Teller sees an abstract scene consisting of clip art objects in a semantically meaningful configuration, while the \Drawer sees a drawing canvas that is initially empty. The goal of the game is to have both players communicate so that the \Drawer can reconstruct the image of the \Teller, without ever seeing it.

Our task requires effective communication because the two players cannot see each other's scenes. The \Teller must describe
the scene in sufficient detail for the Drawer to reconstruct it, which will necessitate grounded language. Moreover, the Drawer will need to carry out a series of actions from a rich action space to position, orient, and resize all of the clip art pieces required for the reconstruction.
Note how clip art pieces form a representation that is perceived visually by humans but is easy to manipulate in a structured manner, in contrast to lower-level pixel-based image representations.
The performance of a pair of agents is judged based on the quality of reconstructed scenes, where high-quality reconstructions result from successful communication.

We collect a {\CoDraw} dataset\footnote{The CoDraw dataset is available at \url{https://github.com/facebookresearch/CoDraw}
} of $\sim$10K variable-length dialogs consisting of $\sim$138K messages with the drawing history at each step of the dialog. We also define a similarity metric for clip art scenes, which allows us to automatically evaluate the effectiveness of agent communication at the end of a dialog and at intermediate states.
We evaluate several \Drawer and \Teller models\footnote{Models are available at \url{https://github.com/facebookresearch/codraw-models}
} automatically as well as by pairing them with humans, and show that long-term planning and contextual reasoning
are key challenges of the \CoDraw task.

As we developed models and protocols for \CoDraw, we found it critical to train the \Teller and the \Drawer separately on disjoint subsets of the training data. Otherwise, the two machine agents may conspire to successfully achieve the goal while communicating using a shared ``codebook'' that bears little resemblance to natural language. We call this separate-training, joint-evaluation protocol \emph{crosstalk}, which prevents learning of mutually agreed upon codebooks, while still checking for goal completion at test time. We highlight crosstalk as one of our contributions, and believe it can be generally applicable to other related tasks~\citep{Sukhbaatar2016,Foerster2016,DeVries2016,Das2017,Lewis2017}.

\section{Related work}

\myparagraph{Language grounded in environments}
Learning language games in a grounded environment has been studied recently~\citep{Wang2016,Wang2017}.
While language in these works is tied to actions that modify the environment, the tasks do not involve multiple agents that need to cooperate.
Other work on grounded instruction following relies on datasets of pre-generated action sequences annotated with human descriptions, rather than using a single end goal~\citep{long2016scone}. Generation models for these tasks are only evaluated based on their ability to describe an action sequence that is given to them~\citep{fried2018unified}, whereas \Teller models for \CoDraw also need to select in a goal-driven manner the action sequence to describe to the \Drawer.
Language grounding has been studied for robot navigation, manipulation, and environment mapping~\citep{tellex2011,Mei15,Daniele16}.
However, these works manually pair each command with robot actions and lack end-to-end training~\citep{tellex2011}, dialog~\citep{Mei15,Daniele16}, or both~\citep{Walter2014}.
Compared to work on navigation~\citep{vogel2010,anderson2018vision,Fried2018SpeakerFollower} where an agent must follow instructions to move itself in a static environment, \CoDraw involves a structured action space for manipulating clip art pieces to form a semantically meaningful configuration.

\myparagraph{End-to-end goal-driven dialog}
Traditional goal-driven agents are often based on `slot filling'~\citep{lemon2006,wang-lemon2013,Yu2015}, in which the structure of the dialog is pre-specified but the individual slots are replaced by relevant information. Recently, end-to-end neural models are also proposed for goal-driven dialog~\citep{Bordes2017,Li2017a,Li2017,he2017}, as well as goal-free dialog or `chit-chat'~\citep{Shang2015,Sordoni2015,Vinyals2015,Li2016,Dodge2016}. Unlike {\CoDraw}, in these approaches, symbols in the dialog are not grounded into visual objects. 

\myparagraph{Emergent communication}
Building on the seminal works by~\citet{Lewis1969,Lewis1975}, a number of recent works study cooperative games between agents where communication protocols emerge as a consequence of training the agents to accomplish shared goals~\citep{Sukhbaatar2016,Foerster2016}.
These methods have typically been applied to learn to communicate small amounts of information, rather than the complete, semantically meaningful scenes used in the \CoDraw task.
In addition, the learned communication protocols are usually not natural~\citep{Kottur2017} or interpretable, whereas the \CoDraw task is designed to develop agents that use human language.

\myparagraph{Language and vision}
The proposed \CoDraw game is related to several well-known language and vision tasks that study grounded language understanding~\citep{karpathy2015deep,donahue2015long,DeVries2016}. 
For instance, in contrast to image captioning~\citep{Vinyals2017,Xu2015,chen2015mind,Lu2016a}, visual question 
answering~\citep{Antol2015,Zhang2015a,Goyal2016,Gao2015,Krishna2016,Malinowski2014,Ren2015,Tapaswi2015,Yu2015,Zhu2015b} and recent embodied extensions~\citep{das2018embodied}, \CoDraw involves multiple rounds of interactions between two agents. 
Both agents hold their own partially observable states and may need to build a model of their partner's state to collaborate effectively.
Compared to past work on generating abstract scenes from single captions \cite{Zitnick2013}, scenes in \CoDraw are reconstructed over multiple rounds, and the task requires \Teller models to generate coherent and precise descriptions over the course of a full dialog.
Compared to visual dialog~\citep{Das2016a,Das2017,Strub2017,Mostafazadeh2017} tasks, agents need to additionally cooperate to change the environment with actions (\eg, move pieces around).
Thus, the agents have to possess the ability to adapt and hold a dialog about partially-constructed scenes that will occur over the course of their interactions.
In addition, we also want to highlight that \CoDraw has a well-defined communication goal, which facilitates objective measurement of success and enables end-to-end goal-driven learning. 

\section{CoDraw task and dataset}

In this section, we first detail our task, then present the \CoDraw dataset, and finally propose a scene similarity metric which allows automatic evaluation of the reconstructed and original scene.

\subsection{Task}

\myparagraph{Abstract scenes} To enable people to easily draw semantically rich scenes on a canvas, we leverage the Abstract Scenes dataset of~\citet{Zitnick2013} and \citet{Zitnick2013a}. This dataset consists of 10,020 semantically consistent scenes created by human annotators.
An example scene is shown in the left portion of Figure~\ref{fig:schema}.
Most scenes contain 6 objects (min 6, max 17, mean 6.67). These scenes depict children playing in a park, and are made from a library of 58 clip arts, including a boy (Mike) and a girl (Jenny) in one of 7 poses and 5 expressions, and various other objects including trees, toys, hats, animals, food, etc. 
An abstract scene is created by dragging and dropping multiple clip art objects to any $(x,y)$ position on the canvas. Spatial transformations can be applied to each clip art, including sizes (small, normal, large) and two orientations (facing left or right).
The clip art serve simultaneously as a high-level visual representation and as a mechanism by which rich drawing actions can be carried out.

\begin{figure*}
\centering
  \centering
  \includegraphics[width=\linewidth,trim=2.9cm 17.0cm 5.0cm 4.3cm,clip]{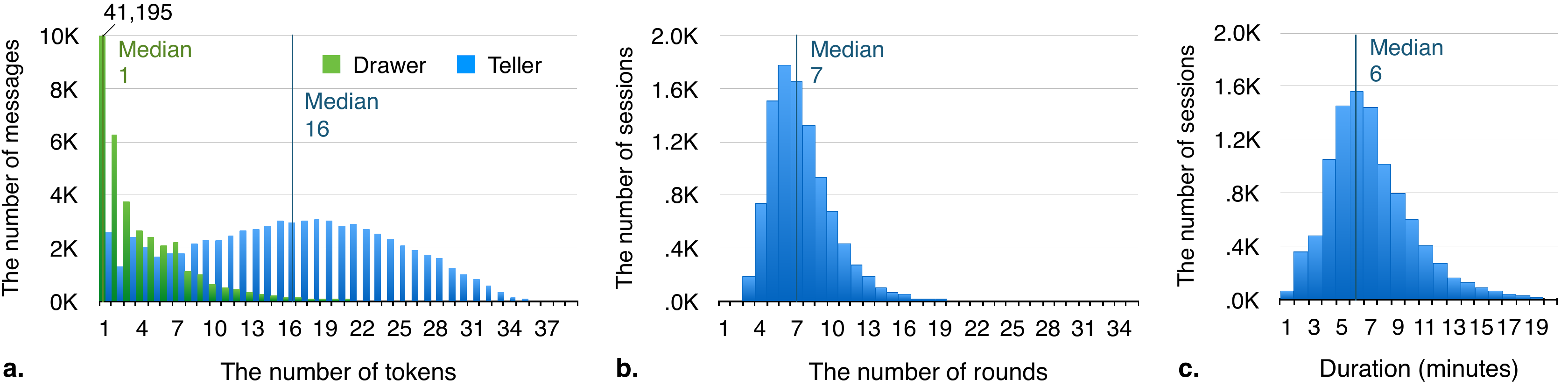}
  \caption{
  Statistics of the \CoDraw~dataset.
  \textbf{(a)} The distribution of the number of tokens in \Teller (blue) and \Drawer (green) messages. Note that the number of single-token messages by {\Drawer}s is 41,195 (62.06\%). The median token counts for {\Teller}s and {\Drawer}s are 16 and 1, respectively.
  \textbf{(b)} The distribution of the numbers of conversation rounds. The median is 7 rounds.
  \textbf{(c)} The distribution of the duration of dialog sessions. The median is 6 minutes.}
  \label{fig:stats}
  \vspace{-0.7em}
\end{figure*}

\myparagraph{Interface} 
We built a drag-and-drop interface based on the Visual Dialog chat interface~\citep{Das2016a} (see Figures~\ref{fig:interface_teller}~and~\ref{fig:interface_drawer} in Appendix~\ref{sec:appendix-interface} for screen shots of the interface). The interface allows real-time interaction between two people. During the conversation, the \Teller describes the scene and answers any questions from the \Drawer on the chat interface, while \Drawer  ``draws'' or reconstructs the scene based on the \Teller's descriptions and instructions. Each side is only allowed to send one message at a time, and must wait for a reply before continuing. The maximum length of a single message is capped at 140 characters: this prevents excessively verbose descriptions and gives the \Drawer more chances to participate in the dialog by encouraging the \Teller to pause more frequently. Both participants were asked to submit the task when they are
both confident that Drawer has accurately reconstructed the
scene of Teller.
To focus the natural language on the high-level semantics of the scene rather than instructions calling for the execution of low-level clip art manipulation actions, the \Teller is not able to observe the \Drawer's canvas while communicating.

\subsection{Dataset}
\label{sec:dataset}

We collect 9,993\footnote{Excluding 27 empty scenes from the original dataset.} dialogs where pairs of people complete the CoDraw task, consisting of one dialog per scene in the Abstract Scenes dataset. The dialogs contain of a total of 138K utterances and include snapshots of the intermediate state of the {\Drawer}'s canvas after each round of each conversation.
See Section~\ref{sec:eval} for a description of how we split the data into training, validation, and test sets.

\myparagraph{Messages}
Figure~\ref{fig:stats}a shows the distribution of message lengths for both {\Drawer}s and {\Teller}s.
The message length distribution for the \Drawer is skewed toward 1 with passive replies like \textit{``ok''}, \textit{``done''}, etc. 
There does exist a heavy tail, which shows that {\Drawer}s ask clarifying questions about the scene like \textit{``where is trunk of second tree, low or high''}.
On the other hand, {\Teller} utterances have a median length of 16 tokens and a vocabulary size of 4,555.
Due to the limited number of clip arts, the vocabulary is smaller than it would be for real images. However, humans still use compositional language to describe clip art configurations and attributes, and make references to previous discourse elements in their messages.

\myparagraph{Rounds} Figure~\ref{fig:stats}b shows the distribution of the numbers of conversational rounds for dialog sessions. Most interactions are shorter than 20 rounds; the median number of rounds is 7.

\myparagraph{Durations} In Figure~\ref{fig:stats}c we see that the median session duration is 6 minutes. We had placed a 20-minute maximum limit on each session.

\subsection{Scene similarity metric}
\label{sec:metric}

The goal-driven nature of
the CoDraw task
naturally lends itself to evaluation by
comparing the reconstructed scene to the original.
For this purpose we define a scene similarity metric, which allows us to automatically evaluate communication effectiveness both at the end of a dialog and at intermediate states. We use the metric to compare how well different machine-machine, human-machine, and human-human pairs can complete the
task.

We represent a scene $C$ as a set of clip art objects $c \in C$, each of which consists of an identifier $id(c)$ that denotes its type, and additional features such as size and $x,y$ position. We denote by $ids(C)$ the set of clip art types that occur in the scene. Given two scenes, the intersection-over-union measure computed over clip art types is:
\begin{equation}
IOU(C, \hat{C}) = \frac{n_\text{int}}{n_\text{union}} = \frac{\sum_{i}\1_{i \in ids(C) \land i \in ids(\hat{C})}}{\left|ids(C) \cup ids(\hat{C})\right|}
\end{equation}
where $n_\text{int}$ ($n_\text{union}$) is the numbers of clip art types in the intersection (union).

To also incorporate features such as size and position, we replace the indicator function in the numerator with a term $g(i, C, \hat{C})$ that measures attribute similarity for shared clip art types. We also introduce a pairwise similarity term $h(i, j, C, \hat{C})$. Overall, scene similarity is defined as:
\begin{equation}
s(C, \hat{C}) = \underbrace{\frac{\sum_i g(i, C, \hat{C})}{n_\text{union}}}_{\text{unary}} + \underbrace{\frac{\sum_{i<j}h(i, j, C, \hat{C})}{n_\text{union}(n_\text{int} - 1)}}_{\text{pairwise}}
\end{equation}

The denominator terms normalize the metric to penalize missing or extra clip art, and we set $g$ and $h$ such that our metric is on a 0-5 scale. The exact terms $g$ and $h$ are described in Appendix~\ref{sec:appendix-metric}.

\section{Models}

We model both the Teller and the Drawer, and evaluate the agents using the metric described in the previous section.
Informed by our analysis of the collected dataset (see Section~\ref{sec:dataset}), we make several modeling assumptions compared to the full generality of the setup that humans were presented with during data collection. These assumptions hold for all models studied in this paper. 

\myparagraph{Assumption 1: Silent \Drawer} We choose to omit the \Drawer's ability to ask clarification questions: our \Drawer models will not generate any messages and our \Teller models will not condition on the text of the \Drawer replies. This is consistent with typical human replies such as \emph{``ok''} or \emph{``done''} (around 62\% of human \Drawer replies only use a single token)  and the fact that the \Drawer talking is not strictly required to resolve the information asymmetry inherent in the task.
We note that this assumption does not reduce the number of modalities needed to solve the task: there is still language generation on the \Teller side, in addition to language understanding, scene perception, and scene generation on the \Drawer side. \Drawer models that can detect when a clarification is required, and then generate a natural language clarification question is interesting future work.


\myparagraph{Assumption 2: Full clip art library} The other assumption is that our drawer models can select from the full clip art library. Humans are only given access to a smaller set so that it can easily fit in the user interface \citep{Zitnick2013a}, while ensuring that all pieces needed to reconstruct the target scene are available. We choose to adopt the full-library condition as the standard for models because it is a stricter evaluation of whether the models are able to make correct grounding decisions.

\subsection{Rule-based nearest-neighbor methods}
\label{sec:baselines}

Simple methods can be quite effective even for what appear to be challenging tasks, so we begin by building models based on nearest-neighbors and rule-based approaches. We split the recorded human conversations available for training into a set of conversation rounds $R$ (possibly from different dialogs), where at each round $r \in R$:
\begin{packed_itemize}
\item \Teller sends a message $m_r$
\item \Drawer removes clip art pieces $C^{(-)}_r$
\item \Drawer adds clip art pieces $C^{(+)}_r$
\item \Drawer replies or ends the conversation
\end{packed_itemize}

\myparagraph{Rule-based nearest-neighbor \Teller}
Our first \Teller model uses a rule-based dialog policy where the \Teller describes exactly one clip art each time it talks. The rule-based system determines which clip art to describe during each round of conversation, following a fixed order that roughly starts with objects in the sky (sun, clouds), followed by objects in the scene (trees, Mike, Jenny), ending with small objects (sunglasses, baseball bat). The message for each object $c$ is then copied from a nearest neighbor in the data:
\begin{align}
R^{(\text{single})} &= \left\{r \in R : C^{(-)}_r = \emptyset, \left\lvert C^{(+)}_r \right\rvert = 1\right\}\\
\hat{r}(c) &= \argmax_{r \in R^{(\text{single})}} s\left(\{c\}, C^{(+)}_r\right) \\
\hat{m}(c) &= m_{\hat{r}(c)}
\end{align}
where $s$ 
is the scene similarity metric from Section~\ref{sec:metric}.
This baseline approach is based on the assumptions that the {\Drawer}'s action was elicited by the \Teller utterance immediately prior, and that the {\Teller}'s utterance will have a similar meaning when copied verbatim into a new conversation and scene context.

\myparagraph{Rule-based nearest-neighbor \Drawer} This \Drawer model is the complement to the rule-based nearest-neighbor \Teller. It likewise follows a fixed rule that the response to each \Teller utterance should be the addition of a single clip art, and uses a character-level string edit distance $d$ to select which clip art object to add to the canvas:
\begin{align}
\hat{r}^\prime(m) &= \argmin_{r \in R^{(\text{single})}} d\left(m, m_r\right) \\
\hat{C}(m) &= C^{(+)}_{\hat{r}^\prime(m)}
\end{align}

\subsection{Neural \Drawer}

\begin{figure*}[t!]
  \centering
  \includegraphics[width=1.0\linewidth]{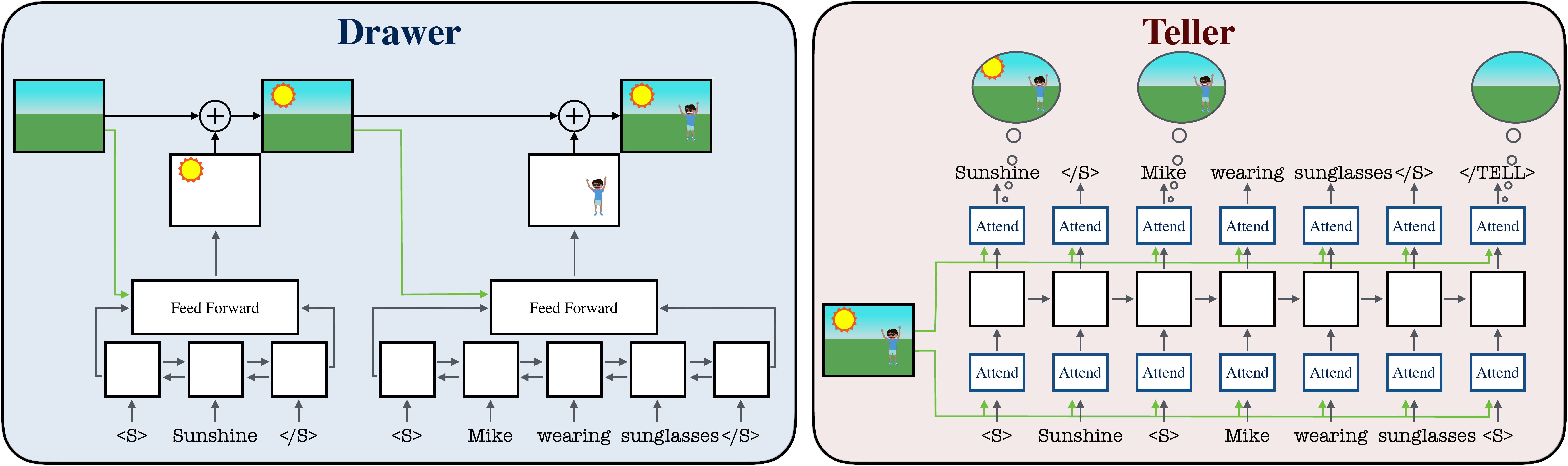}
  \caption{A sketch of our model architectures for the neural Drawer and Teller. The Drawer (left) conditions on the current state of the canvas and a BiLSTM encoding of the previous utterance to decide which clip art pieces to add to a scene. The Teller (right) uses an LSTM language model with attention to the scene (in blue) taking place before and after the LSTM. The ``thought bubbles'' represent intermediate supervision using an auxiliary task of predicting which clip art have not been described yet. In reinforcement learning, the intermediate scenes produced by the drawer are used to calculate rewards. Note that the language used here was constructed for illustrative purposes, and that the messages in our dataset are more detailed and precise.
  }
  \label{fig:model}
  \vspace{-0.7em}
\end{figure*}

Our second \Drawer model is based on the neural network architecture shown in the left portion of Figure~\ref{fig:model}. At each round of conversation, the \Drawer conditions on the {\Teller}'s last message, which is encoded into a vector using a bi-directional LSTM. The \Drawer also uses as input a vector that represents the current state of the canvas. These vectors are then processed by a dense feed-forward neural network to produce a vector that represents the {\Drawer}'s action, which consists of adding a (possibly empty) set of clip art pieces to the drawing. It is trained using a combination of cross-entropy losses (for categorical decisions such as which clip art pieces to add and what orientation to use) and $L_2$ losses that penalizes placing pieces at distant $(x,y)$ coordinates; see Appendix~\ref{sec:appendix-drawer-model} for details.

\subsection{Neural \Teller: scene2seq}

For our neural \Teller models, we adopt an architecture that we call \emph{scene2seq} (right portion of Figure~\ref{fig:model}). This architecture is a conditional language model over the \Teller's side of the conversation with special \emph{next-utterance tokens} to indicate when the \Teller ends its current utterance and waits for a reply from the \Drawer.\footnote{Though none of the models in this paper handle language in the \Drawer replies, these can be incorporated into the \emph{scene2seq} framework similar to the approach of \citet{Lewis2017}.} The language model is implemented using an LSTM, where information about the ground-truth scene is incorporated at both the input and output of each LSTM cell through the use of an attention mechanism. Attention occurs over individual clip art pieces: each clip art in the ground-truth scene is represented using a vector that is the sum of learned embeddings for different clip art attributes (e.g.\ $e_\text{type=Mike}$, $e_\text{size=small}$, etc.) At test time, the \Teller's messages are constructed by decoding from the language model using greedy word selection.

To communicate effectively, the \Teller must keep track of which parts of the scene it has and has not described, and also generate language that is likely to accomplish the task objective when interpreted by the \Drawer. We found that training the \emph{scene2seq} model using a maximum likelihood objective did not result in long-term coherent dialogs for novel scenes. Rather than introducing a new architecture to address these deficiencies, we explore reducing them by using alternative training objectives. To better ensure that the model keeps track of which pieces of information it has already communicated, we take advantage of the availability of drawings at each round of the recorded human dialogs and introduce an auxiliary loss based on predicting these drawings. To select language that 
is more likely to lead to successful task completion, we further fine-tune our \Teller models to directly optimize the end-task goal using reinforcement learning.

\subsubsection{Intermediate supervision}

We incorporate state tracking into the scene2seq architecture through the use of an auxiliary loss. This formulation maintains the end-to-end training procedure and keeps test-time decoding exactly the same; the only change is that during training, at each utterance separator token, the output from the LSTM is used to classify whether each clip art in the ground truth has been drawn already or not. Here we make use of the fact that the CoDraw dataset records human drawer actions at each round of the conversation, not just at the end. The network outputs a score for each clip art type, which is connected to a softmax loss for the clip art in the ground truth scene (the scores for absent clip arts do not contribute to the auxiliary loss). We find that adding such a supervisory signal reduces the \Teller's propensity for repeating itself or omitting objects.

\subsubsection{Reinforcement learning}

The auxiliary loss helps the agent be more coherent throughout the dialog, but it is still an indirect proxy for the end goal of having the \Drawer successfully reconstruct the scene. By training the agents using reinforcement learning (RL), it is possible to more directly optimize for the goal of the task.
In this work we only train the Teller with RL, because the Teller has challenges maintaining a long-term strategy throughout a long dialog, whereas preliminary results showed that making local decisions is less detrimental for {\Drawer}s.
The scene2seq \Teller architecture remains unchanged, and each action from the agent is to output a word or one of two special tokens: a next-utterance token and a stop token. After each next-utterance token, our neural \Drawer model is used to take an action in the scene and the resulting change in scene similarity metric is used as a reward. However, this reward scheme alone has an issue: once all objects in the scene are described, any further messages will not result in a change in the scene and have a reward of zero. As a result, there is no incentive to end the conversation. We address this by applying a penalty of 0.3 to the reward whenever the \Drawer makes no changes to the scene. We train our \Teller with REINFORCE~\citep{williams1992simple}, while the parameters of the \Drawer are held fixed.

\section{Training protocol and evaluation}
\label{sec:eval}

To evaluate our models, we pair our models with other models, as well as with a human.

\myparagraph{Human-machine pairs}
We modified the interface used for data collection to have each trained model to play one game with a human per scene in the test set. We then compare the scene reconstruction quality between human-model pairs for various models and with human-human pairs.

\myparagraph{Script-based \Drawer evaluation}
In addition to human evaluation, we would like to have automated evaluation protocols that can quickly estimate the quality of different models.
\Drawer models can be evaluated against a recorded human conversation from a script (a recorded dialog from the dataset) by measuring scene similarity at the end of the dialog. While this setup does not capture the full interactive nature of the task, the \Drawer model still receives human descriptions of the scene and should be able to reconstruct it. Our modeling assumptions include not giving \Drawer models the ability to ask clarifying questions, which further suggests that script-based evaluation can reasonably measure model quality.

\myparagraph{Machine-machine evaluation}
To evaluate \Teller models in a goal-driven manner, a ``script'' from the dataset is not sufficient. We instead consider an evaluation where a \Teller model and \Drawer model are paired, and their joint performance is evaluated using the scene similarity metric.

\begin{table*}
\begin{center}
\centering
\begin{tabular}{@{}lllc@{}}
\toprule
& Teller & Drawer & Scene similarity \\
\cmidrule[\lightrulewidth]{2-4}
\multirow{3}{*}{\rotatebox[origin=c]{90}{\tiny Script-based} $\begin{dcases} \\ \\ \\ \\[-2.2em] \end{dcases}$}
& Script (replays human messages) & Rule-Based Nearest Neighbor     	 &0.94 \\
& Script (replays human messages) & Neural Network	 &3.39 \\
& Script (replays human messages) & Human & \textbf{3.83} \\
\cmidrule{2-4}
\multirow{4}{*}{\rotatebox[origin=c]{90}{\tiny Human-Machine} $\begin{dcases} \\ \\ \\ \\[-1.2em] \end{dcases}$}
& Rule-based Nearest Neighbor  & Human &      	 3.21 \\
& Scene2seq (imitation learning) & Human &      	 2.69 \\
& \hspace{1em}+ auxiliary loss & Human &      	 3.04 \\
& \hspace{2em}+ RL fine-tuning  & Human & \textbf{3.65} \\
\cmidrule{2-4}
\multirow{4}{*}{\rotatebox[origin=c]{90}{\tiny Machine-Machine} $\begin{dcases} \\ \\ \\ \\[-1.2em] \end{dcases}$}  
& Rule-based Nearest Neighbor  &     	 Neural Network &	 3.08 \\
& Scene2seq (imitation learning)  & Neural Network &	 2.67 \\
& \hspace{1em}+ auxiliary loss  & Neural Network &	 3.02 \\
& \hspace{2em}+ RL fine-tuning  &   Neural Network &	 \textbf{3.67} \\
\cmidrule{2-4}
& Human & Human & \textbf{4.17} \\
\bottomrule
\end{tabular}
\end{center}
\caption{\label{tab:results}Results for our models on the test set, using three types of evaluation: script-based (i.e. replaying \Teller utterances from the dataset), human-machine, and machine-machine pair evaluation.}
\end{table*}

\subsection{Crosstalk training protocol}

\begin{figure}
  \centering
  \adjincludegraphics[width=0.9\columnwidth,trim={0 0 {0.515\width} 0},clip]{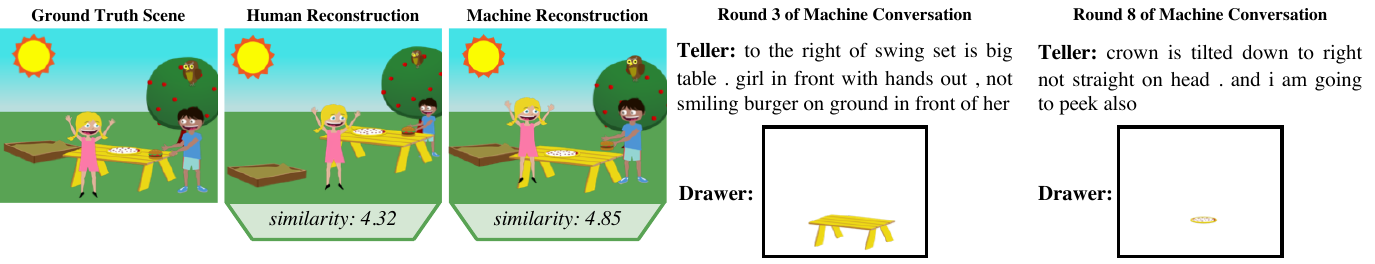}
  \adjincludegraphics[width=\columnwidth,trim={{0.486\width} 0 0 0},clip]{figures/qualitative_codebook2}
  \caption{A rule-based nearest-neighbor {\Teller} and {\Drawer} pair ``trained" on the same data outperforms humans for this scene according to the similarity metric, but the language used by the models doesn't always correspond in meaning to the actions taken. The top row shows a scene from the test set and corresponding human/model reconstructions. The bottom row shows the \Teller message and \Drawer action from two rounds of conversation by the machine agents.}
  \label{fig:codebook}
\end{figure}

Automatically evaluating agents, especially in the machine-machine paired setting, requires some care because a pair of agents can achieve a perfect score while communicating in a shared code that bears no resemblance to natural language. There are several ways such co-adaptation can develop.
One is by overfitting to the training data to the extent that it's used as a codebook -- we see this with the rule-based nearest-neighbor agents described in Section~\ref{sec:baselines}, where a \Drawer-\Teller pair ``trained'' on the same data outperforms humans on the CoDraw task.
An examination of the language, however, reveals that only limited generalization has taken place (see Figure~\ref{fig:codebook}). Another way that agents can co-adapt is if they are trained jointly, for example using reinforcement learning.
To limit these sources of co-adaptation, we propose a training protocol we call ``crosstalk.'' In this setting, the training data is split in half, and the \Teller and \Drawer are trained separately on disjoint halves of the training data.
When joint training of a {\Teller}-{\Drawer} pair is required (as with reinforcement learning), the training process is run separately for both halves of the training data, but evaluation pairs a \Teller trained on the first partition with a \Drawer trained on the second.
This ensures that models can succeed only by communicating in a way that generalizes to new conversation partners, and not via a highly specialized codebook specific to model instances.

Taking the crosstalk training protocol into account, the dataset split we use for all experiments is: 40\% \Teller training data (3,994 scenes/dialogs), 40\% \Drawer training data (3,995), 10\% development data (1,002) and 10\% testing data (1,002).

\section{Results}
\label{sec:results}

Results for our models are shown in Table~\ref{tab:results}. All numbers are scene similarities, averaged across scenes in the test set.

\myparagraph{Neural \Drawer is the best \Drawer model}
In the script setting, our neural \Drawer is able to outperform the rule-based nearest-neighbor baseline (3.39 vs.\ 0.94) and close most of the gap between baseline (0.94) and human performance (4.17).

\myparagraph{Validity of script-based \Drawer evaluation} To test the validity of script-based \Drawer evaluation -- where a \Drawer is paired with a \Teller that recites the human script from the dataset corresponding to the test scenes -- we include results from interactively pairing human {\Drawer}s with a \Teller that recites the scripted messages.
While average scene similarity is lower than when using live human {\Teller}s (3.83 vs. 4.17), the scripts are sufficient to achieve over 91\% of the effectiveness of the same \Teller utterances when they were communicated live (according to our metric).
The drop in similarity may be in part because the \Teller can't answer clarifying questions specific to the {\Drawer}'s personal understanding of the instructions.
Note that a human \Drawer with a script-based \Teller still outperforms our best \Drawer model paired with a script-based \Teller.

\myparagraph{Benefits of intermediate supervision and goal-driven training} Pairing our models with humans shows that the \emph{scene2seq} \Teller model trained with imitation learning is worse than the rule-based nearest-neighbor baseline (2.69 vs.\ 3.21), but that the addition of an auxiliary loss followed by fine-tuning with reinforcement learning allow it to outperform the baseline (3.65 vs.\ 3.21). However, there is still a gap compared to human {\Teller}s (3.65 vs.\ 4.17). Many participants in our human study noted that they received unclear instructions from the models they were paired with, or expressed frustration that their partners could not answer clarifying questions as a way of resolving such situations. Recall that our \Teller models currently ignore any utterances from the \Drawer.

\myparagraph{Correlation between fully-automated and human-machine evaluation} We also report the result of paired evaluation for different \Teller models and our best \Drawer, showing that the relative rankings of the different \Teller types match those we see when models are paired with humans. This shows that automated evaluation while following the crosstalk training protocol is a suitable automated proxy for human-evaluation.

\subsection{Typical errors}

The errors made by \Teller reflect two key challenges posed by the CoDraw task: reasoning about the context of the conversation and what has already been drawn so far, and planning ahead to fully and effectively communicate the required information. A common mistake the rule-based nearest-neighbor \Teller makes is to reference objects that are not present in the current scene. Figure~\ref{fig:codebook} shows an example (bottom left) where the \Teller has copied a message referencing a ``swing'' that does not exist in the current scene. In a sample of 5 scenes from the test set, the rule-based nearest-neighbor \Teller describes a non-existent object 11 times, compared to just 1 time for the scene2seq \Teller trained with imitation learning. The scene2seq \Teller, on the other hand, frequently describes clip art pieces multiple times or forgets to mention some of them: in the same sample of scenes, it re-describes an object 10 times (vs.\ 2 for the baseline) and fails to mention 11 objects (vs.\ 2.) The addition of an auxiliary loss and RL fine-tuning reduces these classes of errors while avoiding frequent descriptions of irrelevant objects (0 references to non-existent objects, 3 instances of re-describing an object, and 4 objects omitted.)

On the \Drawer side, the most salient class of mistakes made by the neural network model is semantically inconsistent placement of multiple clip art pieces. Several instances of this can be seen in Figure~\ref{fig:appendix-qualitative-script} in Appendix~\ref{sec:appendix-qualitative}, where the \Drawer places a hat in the air instead of on a person's head, or where the drawn clip art pieces overlap in a visually unnatural way.

Qualitative examples of both human and model behavior are provided in Appendix~\ref{sec:appendix-qualitative}.

\section{Conclusion}

In this paper, we introduce CoDraw: a collaborative task designed to facilitate learning of effective natural language communication in a grounded context. The task combines language, perception, and actions while permitting automated goal-driven evaluation both at the end and as a measure of intermediate progress. We introduce a dataset and models for this task, and propose a \emph{crosstalk} training + evaluation protocol that is more generally applicable to studying emergent communication. The models we present in this paper show levels of task performance that are still far from what humans can achieve. Long-term planning and contextual reasoning as two key challenges for this task that our models only begin to address. We hope that the grounded, goal-driven communication setting that CoDraw is a testbed for can lead to future progress in building agents that can speak more naturally and better maintain coherency over a long dialog, while being grounded in perception and actions. 

\subsubsection*{Acknowledgments}

We thank C. Lawrence Zitnick for helpful comments and discussion.
Byoung-Tak Zhang was partly supported by the Institute for Information \& Communications Technology Promotion (R0126-16-1072-SW.StarLab, 2017-0-01772-VTT) grant funded by the Korea government.

\bibliography{acl2019}
\bibliographystyle{acl_natbib}

\clearpage
\appendix
\onecolumn
\section{Interface and data collection}
\label{sec:appendix-interface}

\subsection{Interface}

\begin{figure}[h]
\centering
  \centering
  \includegraphics[width=.7\linewidth]{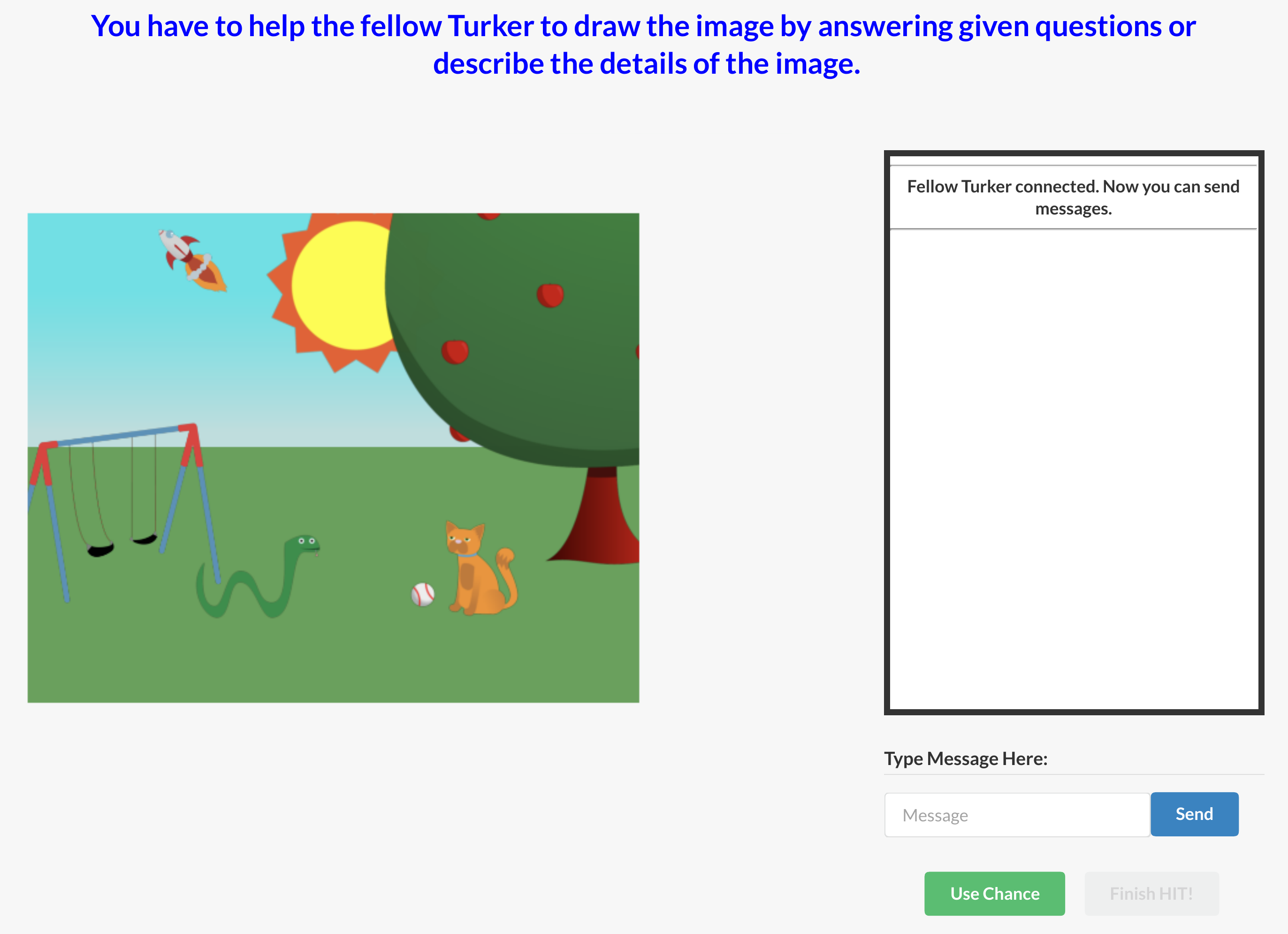}
  \caption{User interface for \Teller. The left image is an abstract scene from \citet{Zitnick2013a}. The \Teller sends messages using an input box. The \Teller has a single chance to peek at the \Drawer's canvas to correct mistakes. The \Teller can decide when to finish the session.}
  \label{fig:interface_teller}
\end{figure}

\begin{figure}[h]
  \centering
  \includegraphics[width=.7\linewidth]{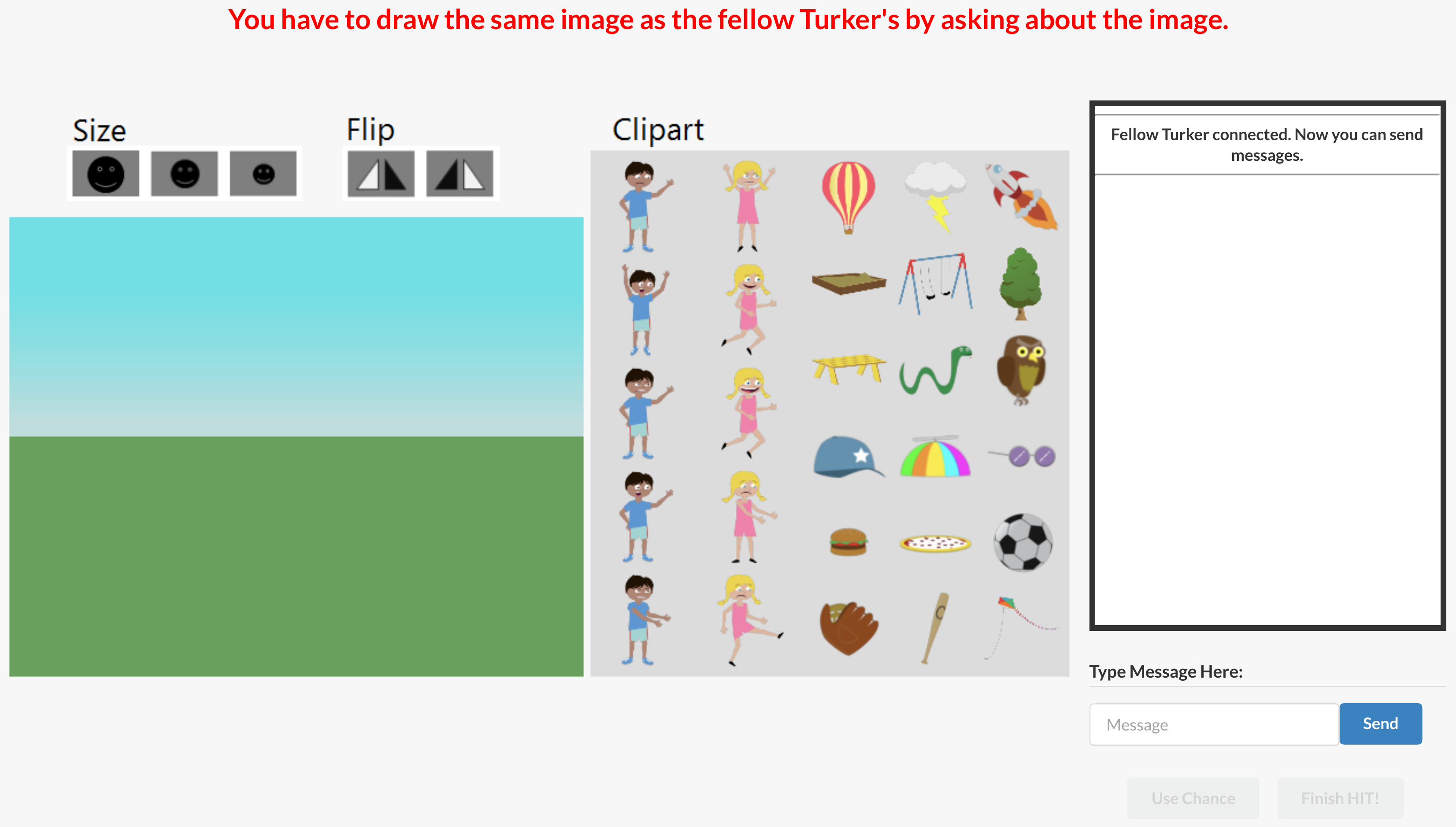}
  \caption{User interface for a \Drawer. The \Drawer has an empty canvas and a randomly generated drawing palette of Mike, Jenny, and 18 other objects, chosen from a library of 58 clip arts. We ensure that using the available objects, the \Drawer can fully reproduce the scene. Using the library, the \Drawer can draw on the canvas in a drag-and-drop fashion. The \Drawer can also send messages using the provided input box. However, the peek button is disabled: only the \Teller can use it.}
  \label{fig:interface_drawer}
\end{figure}

Figure~\ref{fig:interface_teller} shows the interface for the \Teller, and Figure~\ref{fig:interface_drawer} shows the interface for the \Drawer. Following previous works~\citep{Zitnick2013,Zitnick2013a}, {\Drawer}s are given 20 clip art objects selected randomly from the 58 clip art objects in the library, while ensuring that all objects required to reconstruct the scene are available. 


\subsection{Additional interaction: a chance to peek}

To make sure that the natural language focused on the high-level semantics of the scene rather than instructions calling for the execution of low-level clip art manipulation actions, we did not allow \Teller to continuously observe \Drawer's canvas. However, direct visual feedback may be necessary to get the all the details right. We also hypothesize that such feedback would help human participants calibrate themselves when they are new to the task (models do not have this issue because of the rich supervisory signal available in the collected dataset.)

To capture this idea, we give one chance for the \Teller to look at the \Drawer's canvas using a `peek' button in the interface. Communication is only allowed after the peek window is closed.

Although we consider the ability to peek to be a part of the CoDraw task, we leave for future work the creation of models that can strategically reason about when to use this chance in a way that maximizes task effectiveness. We note that omitting this behavior from the \Teller models described in this paper does not decrease the number of modalities needed to complete the task -- our models still incorporate language understanding, language generation, perception, and action.

\subsection{Participant statistics}
We found that approximately 13.6\% of human participants disconnected early, prior to fully completing the task with their partner. We paid participants who stayed in the conversation and had posted at least three messages. However, we exclude those incomplete sessions in the dataset, and only use the completed sessions.

There are 616 unique participants represented in our collected data. Among these workers, the 5 most active have done 26.63\% of all finished tasks (1,419, 1,358, 1,112, 1,110, and 1,068 tasks). Across all workers, the maximum, median, and minimum numbers of tasks finished by a worker are 1,419, 3, and 1, respectively.

\subsection{Pre-processing}

We pre-process all collected \Teller and \Drawer utterances using the Bing Spell Check API\footnote{\url{https://www.microsoft.com/cognitive-services/en-us/bing-spell-check-api}}. The text is then tokenized using the Python Natural Language Toolkit, \texttt{nltk} \citep{Bird2009}. We release the token sequences after pre-processing as part of the CoDraw dataset, so that different models may be compared in a standard set of data conditions. At the same time, raw (unprocessed) text is also made available, to allow revisiting the pre-processing decisions should the need arise.

\newpage

\begin{figure}
\centering
  \centering
  \begin{subfigure}[b]{0.32\textwidth}
  \centering
  \includegraphics[width=\textwidth]{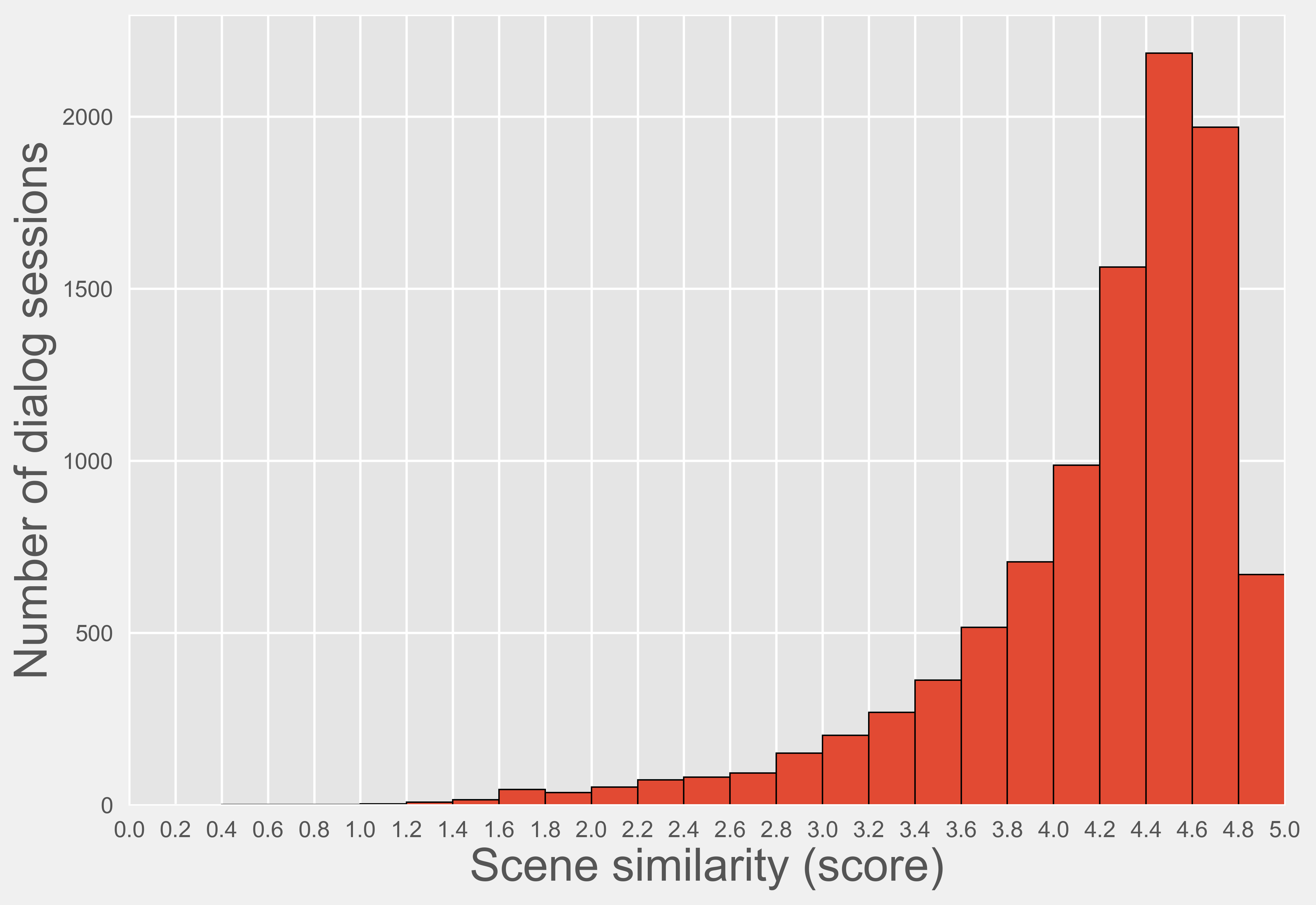}
  \caption{}
  \label{fig:stat-score}
  \end{subfigure}
  \begin{subfigure}[b]{0.32\textwidth}
  \centering
  \includegraphics[width=\textwidth]{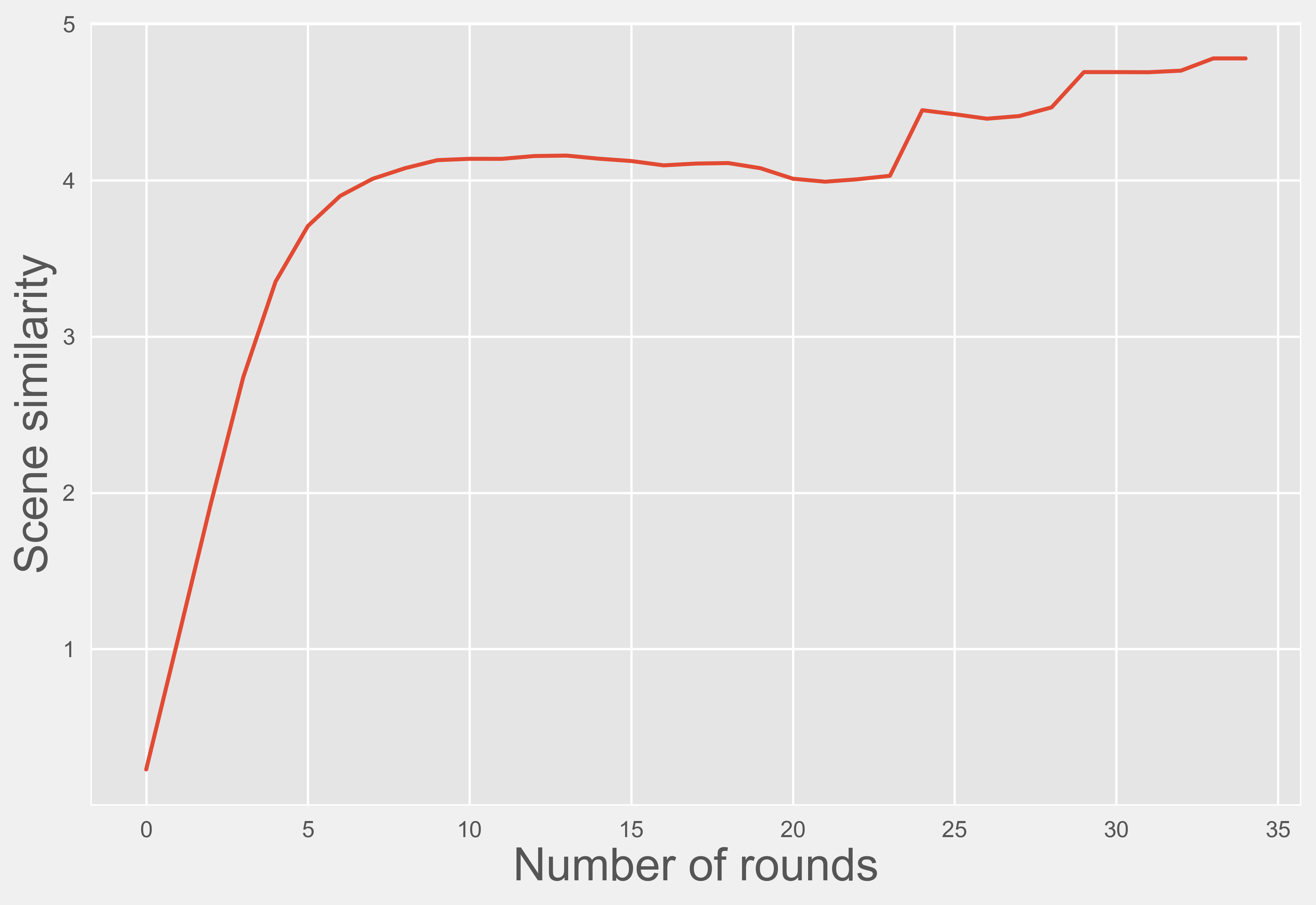}
  \caption{}
  \label{fig:similarity-over-time-1}
  \end{subfigure}
  \begin{subfigure}[b]{0.32\textwidth}
  \centering
  \includegraphics[width=\textwidth]{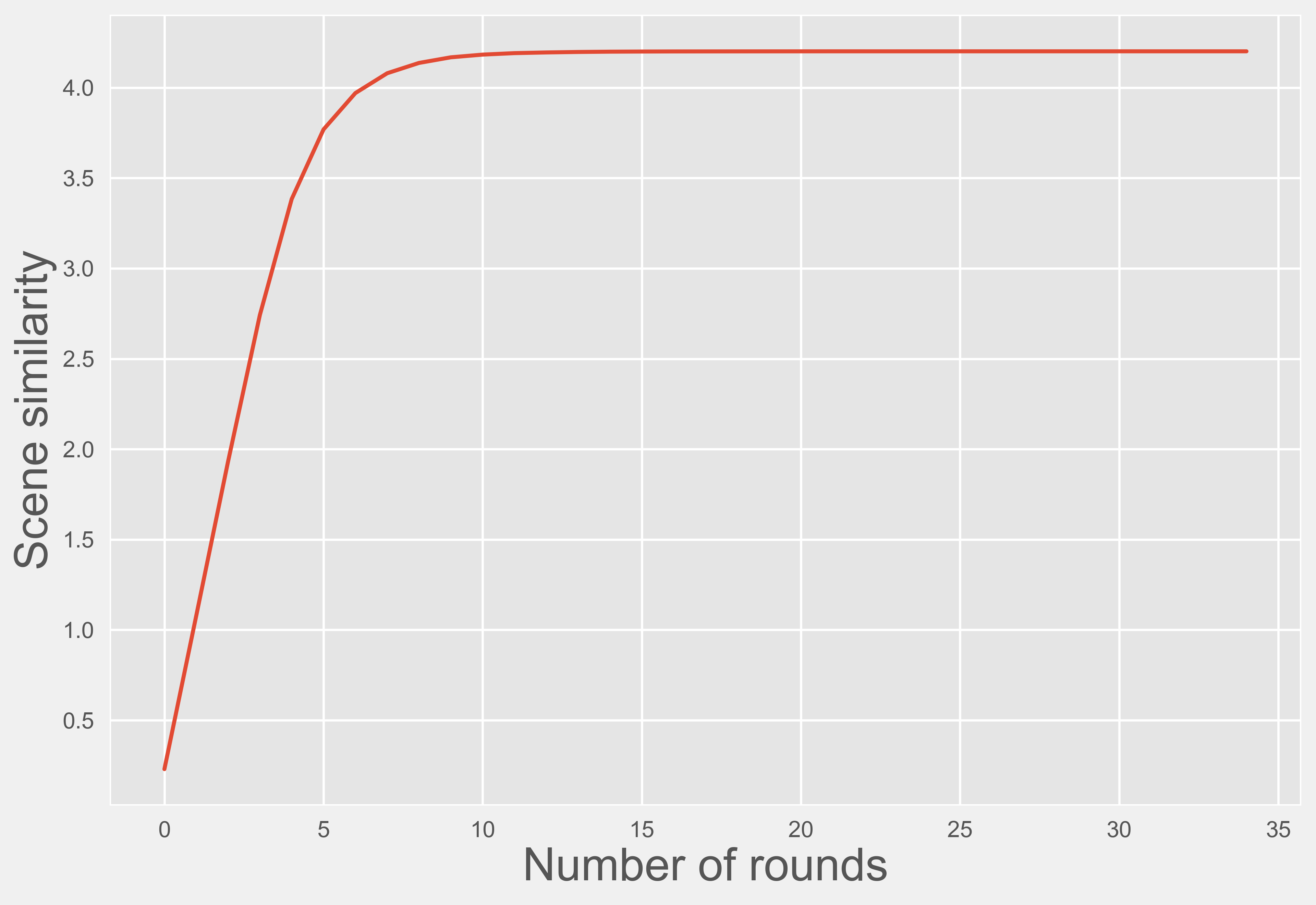}
  \caption{}
  \label{fig:similarity-over-time-2}
  \end{subfigure}
  \caption{\textbf{(a)} The distribution of overall scores at the end of the dialog. \textbf{(b-c)} Average scene similarity plotted for different conversation rounds. In (b), only conversations that have reached the given number of rounds are included. In (c), conversations that end early are padded to 35 rounds through the addition of empty messages/actions.}
\end{figure}

\section{Scene similarity metric}
\label{sec:appendix-metric}

The clip art library consists of $58$ base clip art types  (e.g.\ the sun, a cloud, Mike, Jenny, soccer ball, etc.) Each clip art object $c$ consists of an identifier $id(c)$ that denotes its type, an indicator feature vector $f(c)$ that determines properties such such as size and orientation (e.g.\ $\1_{\text{size}=\text{small}}$, $\1_{\text{size}=\text{medium}}$, etc.\ for a total of 41 binary features), and two real-valued features $x(c)$ and $y(c)$ that encode the $x$ and $y$ position on the canvas, normalized to the 0-1 range.

We represent a scene $C$ as a set of individual clip art objects $c \in C$. We denote by $ids(C)$ the set of clip art types that occurs in the scene. Following \citet{Zitnick2013}, a given clip art type may occur at most once in the scene; let $C[i]$ be the clip art $c \in C$ such that $id(c) = i$.

Given a ground-truth scene $C$ and a predicted scene $\hat{C}$ scene similarity $s$ is defined as:
$$s(C, \hat{C}) = \underbrace{\frac{\sum_{i \in ids(C) \cap ids(\hat{C})}g(C[i], \hat{C}[i])}{|ids(C) \cup ids(\hat{C})|}}_{\text{unary}} + \underbrace{\frac{\sum_{i, j \in ids(C) \cap ids(\hat{C}), i<j} h(C[i], C[j], \hat{C}[i], \hat{C}[j])}{|ids(C) \cup ids(\hat{C})|(|ids(C) \cap ids(\hat{C})| - 1)}}_{\text{pairwise}}$$
where
\begin{align*}
g(c, \hat{c}) = &w_0 \\
&- w_1 \1_\text{clip art piece $\hat{c}$ faces the wrong direction} \\
&- w_2 \1_\text{clip art piece  $\hat{c}$ is Mike or Jenny and has the wrong facial expression} \\
&- w_3 \1_\text{clip art piece  $\hat{c}$ is Mike or Jenny and has the wrong body pose} \\
&- w_4 \1_\text{clip art piece  $\hat{c}$ has the wrong size} \\
&- w_5 \sqrt{\left({x(\hat{c}) - x(c)}\right)^2 + \left({y(\hat{c}) - y(c)}\right)^2} \\
\\
h(c_i, c_j, \hat{c}_i, \hat{c}_j) = &- w_6 1_{(\hat{x}_{c_i} - \hat{x}_{c_j})(x_{c_i} - x_{c_j}) < 0} \\
&- w_7 1_{(\hat{y}_{c_i} - \hat{y}_{c_j})(y_{c_i} - y_{c_j}) < 0} \\
\end{align*}

We use parameters $\vw = [5, 1, 0.5, 0.5, 1, 1, 1, 1]$, which provides a balance between the different components and ensures that scene similarities are constrained to be between 0 and 5.

Figure~\ref{fig:stat-score} shows the distribution of scene similarity scores throughout the dataset. Figure~\ref{fig:similarity-over-time-1}-\subref{fig:similarity-over-time-2} shows the progress of scene similarity scores over the rounds of a conversation. An average conversation is done improving the scene similarity after about 5 rounds, but for longer conversations that continue to 23 rounds, there is still room for improvement.

\newpage
\section{Neural \Drawer architecture}
\label{sec:appendix-drawer-model}

In this section, we describe in greater detail our neural network architecture approach for the \Drawer. Contextual reasoning is an important part of the CoDraw task: each message from the \Teller can relate back to what the \Drawer has previously heard or drawn, and the clip art pieces it places on the canvas must form a semantically coherent scene. To capture these effects, our model should condition on the past history of the conversation and use an action representation that is conducive to generating coherent scenes.

When considering past history, we make the Markovian assumption that the current state of the \Drawer's canvas captures all information from the previous rounds of dialog. Thus, the \Drawer need only consider the most recent utterance from the \Teller and the current canvas to decide what to draw next. 
We experimented with incorporating additional context -- such as previous messages from the \Teller or the action sequence by which the \Drawer arrived at its current canvas configuration -- but did not observe any gains in performance.

We represent the state of the canvas with a vector $v_{canvas}$ that is the concatenation of feature vectors for each of the 58 possible clip art types:
$$v_{canvas}(C) = \left[v_0(C) ; v_1(C) ; \ldots ; v_{57}(C)\right]$$
$$\text{where } v_i(C) = \begin{cases} \left[1 ; f(C[i]) ; x(C[i]) ; y(C[i])\right] &\mbox{if } i \in ids(C) \\
\mathbf{0} & \mbox{otherwise}\end{cases} $$
The individual feature vectors $v_i(C)$ represent binary and $(x,y)$ features of the clip art piece if it is present on the canvas, and are zeroed out if a clip art of the given type is not present on the canvas.

The most recent \Teller utterance is encoded into a vector $v_{msg}$ using a bi-directional LSTM. A vector representing the {\Drawer}'s action is then computed using a feed-forward network with a rectified linear unit (ReLU) nonlinearity:
$$v_{action} = W_{out} relu(W_{canvas}v_{canvas} + W_{msg}v_{msg} + b_{in}) + b_{out}$$

The action representation $v_{action}$ has the form:
$$v_{action} = \left[a_0 ; a_1 ; \ldots ; a_{57}\right]$$
\[\text{where } a_i =
\begin{bmatrix}
q(i \in ids(C)) \\
q(f_0(C[i]) = 1 | i \in ids(C)) \\
q(f_1(C[i]) = 1 | i \in ids(C)) \\
\ldots \\
\hat{x}(C[i]) \\
\hat{y}(C[i]) \\
\end{bmatrix}
\]
The values $\hat{x}(C[i])$ and $\hat{y}(C[i])$ are the predicted location for clipart $C[i]$ if it is placed on the canvas, and each quantity $q(event)$ is a logit corresponding to a particular event. The probability of adding a clip art piece to the scene is calculated using the sigmoid function:
$$p(i \in ids(C)) = \frac{1}{1 + \exp -q(i \in ids(C))}$$
while all other probabilities are calculated by applying softmax to each set of mutually-exclusive outcomes, e.g.:
$$p(size(C[i]) = \text{small} | i \in ids(C)) = \frac{\exp{\left(q(size(C[i]) = \text{small} | i \in ids(C))\right)}}{\sum_{s \in \left\{ \text{small}, \text{medium}, \text{large} \right\} } \exp{\left(q(size(C[i]) = s | i \in ids(C))\right)}}$$

At inference time, the {\Drawer}'s action is chosen using greedy decoding. A clip art of type $i$ is added to the canvas if $p(i \in ids(C)) > 0.5$, in which case it is placed at location $(\hat{x}(C[i]), \hat{y}(C[i]))$ with its orientation, size, and other attributes set to their most probable values (as determined by the vector $a_i$.) 

The model is trained using a combination of cross-entropy losses (that maximize the probability of the categorical decisions present in the human action) and an $L_2$ loss that compares the locations where the human placed each clip art piece with the model's estimate.


\section{Qualitative examples}
\label{sec:appendix-qualitative}

\begin{figure}
  \centering
  \includegraphics[width=0.95\columnwidth]{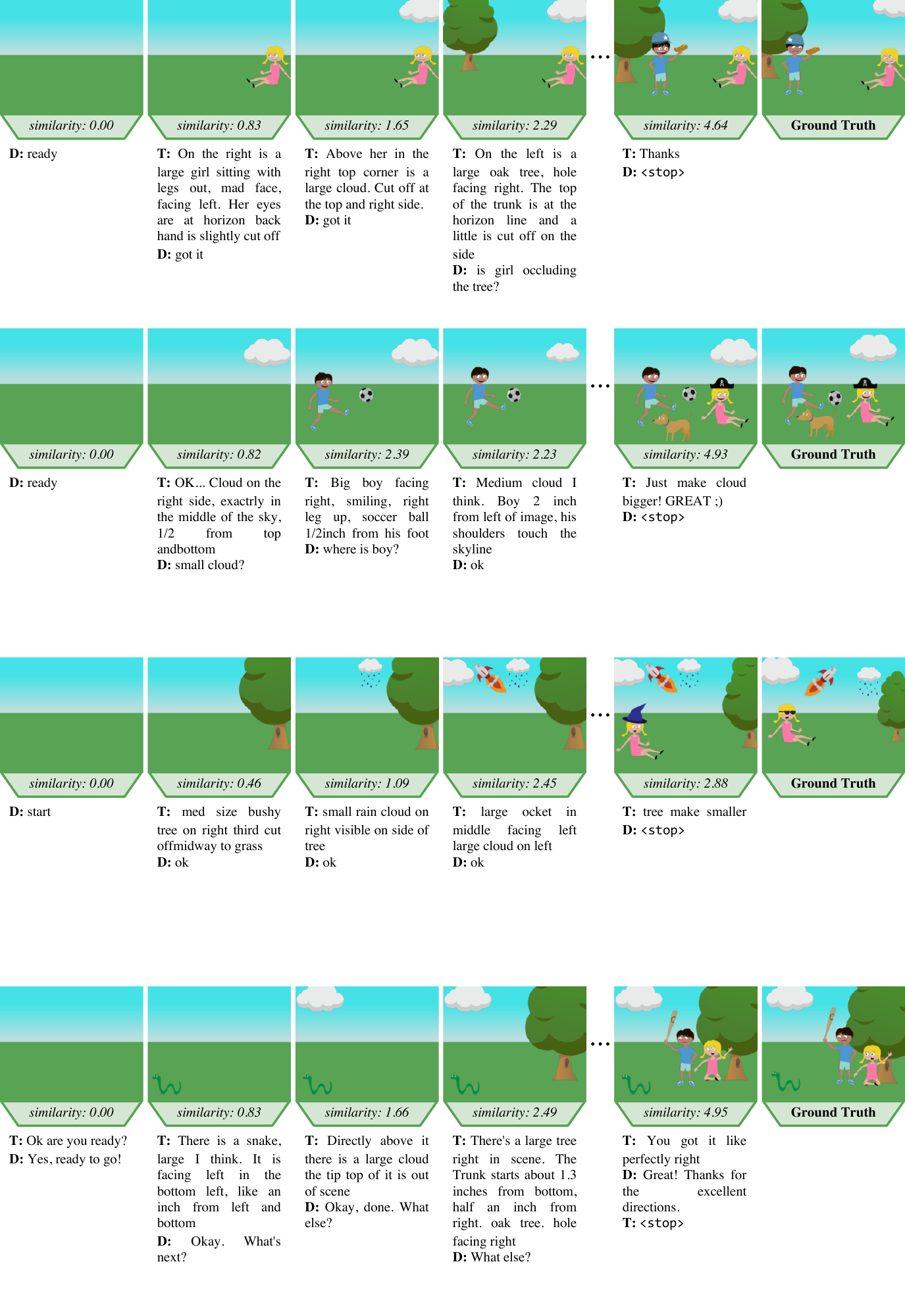}
  \caption{Examples from the Collaborative Drawing (CoDraw) dataset, chosen at random from the test set. The images depict the {\Drawer}'s canvas after each round of conversation. From left to right, we show rounds one through four, then the last round, followed by the ground truth scene. The corresponding conversations between the \Teller (T) and \Drawer (D) are shown below the images. Note that there is no restriction on which of the two participants begins or ends the dialog.}
  \label{fig:appendix-qualitative-human}

\end{figure}

\begin{figure}
  \centering
  \includegraphics[width=0.95\columnwidth]{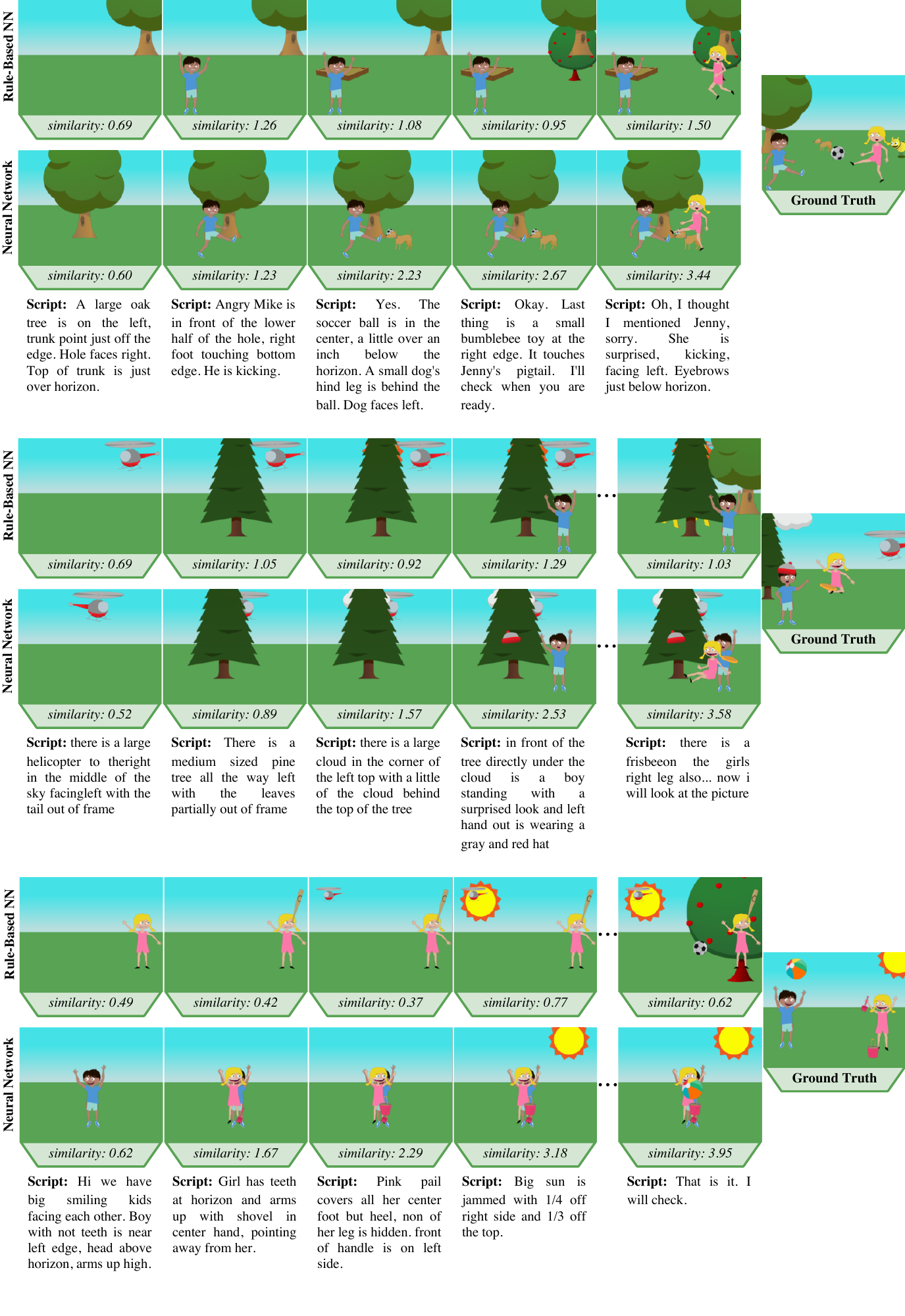}
  \caption{\Drawer model behavior where the \Teller is simulated by replaying messages associated with three randomly-selected scenes from the test set. The images depict the {\Drawer}'s canvas after each round of conversation. For each of the three scenes, the behavior of the Rule-Based Nearest-Neighbor \Drawer is shown in the upper row and the behavior of the Neural Network \Drawer is shown in the lower row.}
  \label{fig:appendix-qualitative-script}
\end{figure}

\begin{figure}
  \centering
  \includegraphics[width=0.95\columnwidth,trim={0 5.911pt 0 0},clip]{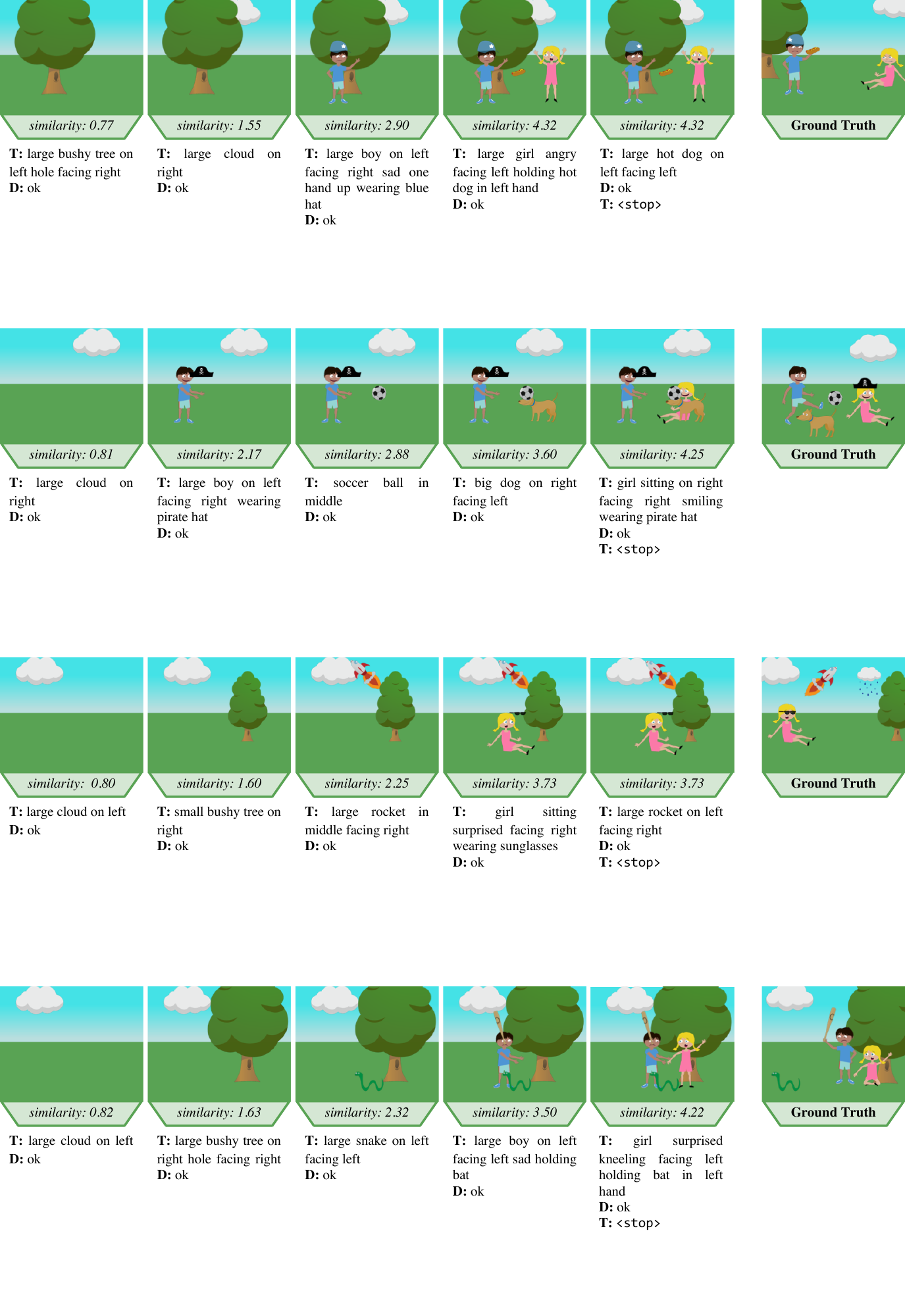}
  \caption{Dialogs from our best \Teller model (scene2seq with an auxiliary loss and RL fine-tuning) communicating with our best \Drawer model (Neural Network \Drawer). The dialogs feature the same scenes as in Figure~\ref{fig:appendix-qualitative-human}, which were sampled at random from the test set.
From left to right, we show the first to the fifth rounds of conversations, followed by the ground truth scene. Our \Teller model chose to use exactly five rounds for each of these four scenes. The corresponding conversations between \Teller (T) and \Drawer (D) are shown below the images.
}
    \label{fig:appendix-qualitative-machine}
\end{figure}

Figure~\ref{fig:appendix-qualitative-human} shows some examples of scenes and dialogs from the CoDraw dataset. The behavior of our \Drawer and \Teller models on a few randomly-selected scenes is illustrated in Figures~\ref{fig:appendix-qualitative-script},~\ref{fig:appendix-qualitative-tellers},~and~\ref{fig:appendix-qualitative-machine}. 

\newcommand{\tellermsg}[1]{#1}

\begin{figure}
  \centering
  \tiny
   {\normalsize Scene A} \hspace{1em} \raisebox{-.5\height}{\includegraphics[width=0.2\linewidth]{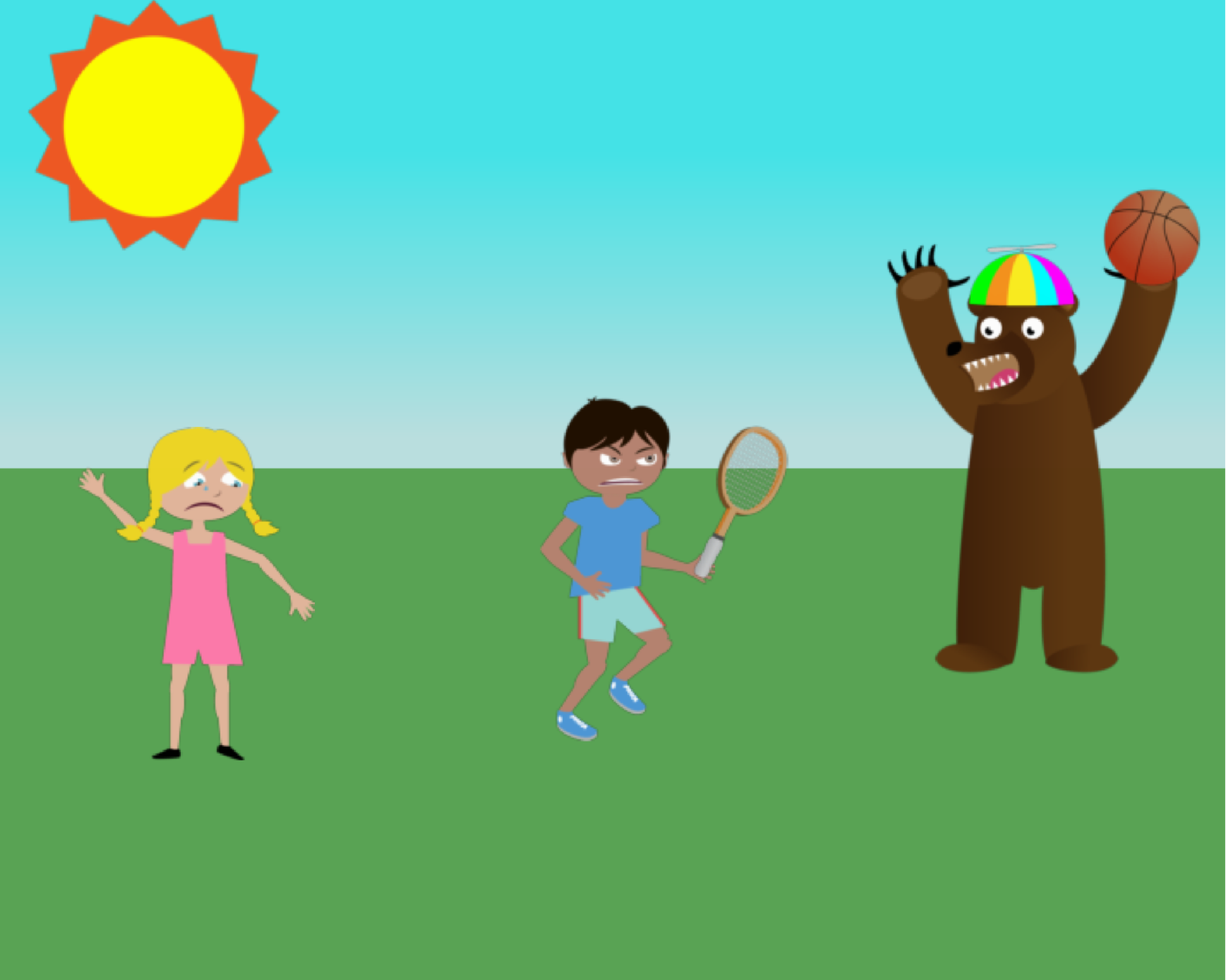}}
  \\[1em]
  \begin{tabular}{@{}p{2em}p{0.2\linewidth}p{0.2\linewidth}p{0.2\linewidth}p{0.2\linewidth}@{}}
  \toprule
   \multirow{2}{*}{\makecell[lb]{Scene,\\ Round}} &\multirow{2}{*}{\makecell[lb]{Rule-Based\\ Nearest-Neighbor}} & \multicolumn{3}{c}{Scene2seq} \\
   \cmidrule{3-5}
  && Imitation learning & + Auxiliary Loss & + RL Fine-tuning \\
  \midrule
  A1 &\tellermsg{in the top left corner is a medium sun very close to the top and left edge}
&\tellermsg{medium sun on left , 1 2 inch from top and side}&\tellermsg{on the right hand side is a medium bear facing left with a basketball in his left hand}
&\tellermsg{medium sun top left}
\\
  \cmidrule{1-5}
A2 &\tellermsg{guy angry arms down facing right like his right knees touches the top right corner of the table}
&\tellermsg{medium bear on right , facing left , 1 2 '' from bottom , small basketball in left hand , racket in left hand}
&\tellermsg{in the middle of the grass is a medium sized boy facing right with a racket in his left hand}
&\tellermsg{medium boy on right angry facing right holding racket}
\\
\cmidrule{1-5}
A3&\tellermsg{right under it is medium girl , waving with a sad face , facing right .}
&\tellermsg{small bear on right side , facing left , 1 2 '' from bottom , small bear on left facing right , armpits on horizon ,}
&\tellermsg{he is holding a racket in his left hand and a tennis racket in the middle of the sky}
&\tellermsg{medium girl on left facing right sad one arm up holding basketball}
\\
  \cmidrule{1-5}
  A4
  &\tellermsg{on the right side of screen , medium bear , chest on horizon , paw almost touching right side of screen}
&\tellermsg{small bear on right facing left , chest at horizon , small basketball in left hand , 1 inch from left , 1 2 from bottom}
&\tellermsg{on the left hand side is a medium girl with sad face facing right with a racket in her right hand .}
&\tellermsg{medium bear on right facing left wearing rainbow hat basketball racket in right hand}
\\
  \cmidrule{1-5}
A5
&\tellermsg{he also has a rainbow hat , but it 's not snugly on him , it 's only slightly touching his head .}
&\tellermsg{small bear on right facing left , chest at horizon , small basketball in left hand , racket in left hand , racket in left hand , racket in left}
&\tellermsg{the bear is facing left . the bear is facing right . the bear is facing left .}
& \texttt{<STOP>}\\
  \cmidrule{1-5}
A6&\tellermsg{sorry there 's a basketball above the boy 's head near the tree}&\texttt{<STOP>}&\texttt{<STOP>}
\\
  \cmidrule{1-5}
A7&\tellermsg{she is holding a racket in her left hand .} \\
  \cmidrule{1-5}
A8&\texttt{<STOP>}
\\
  \bottomrule
  \end{tabular}
  
  \vspace{1em}
  {\normalsize Scene B} \hspace{1em} \raisebox{-.5\height}{\includegraphics[width=0.2\linewidth]{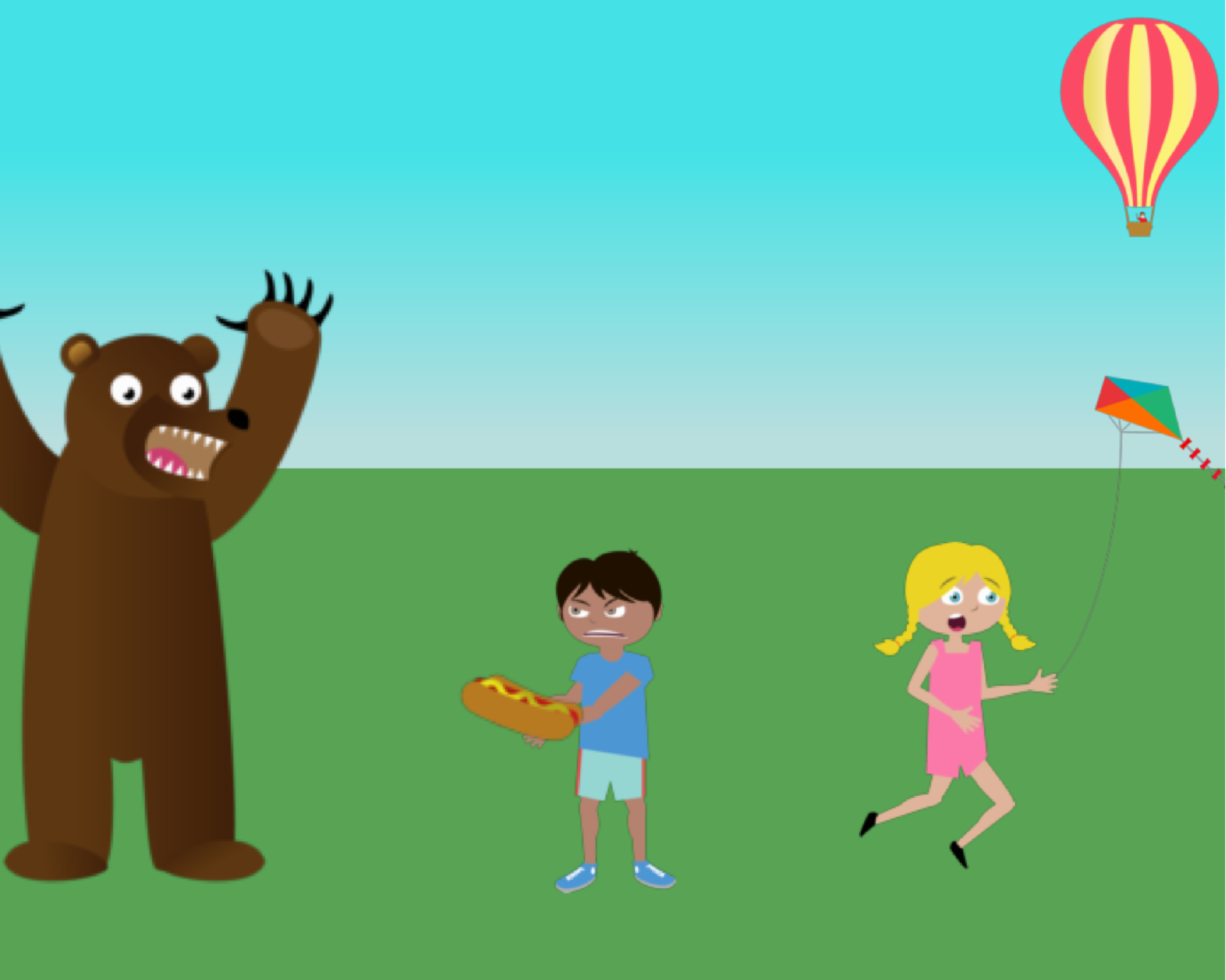}}
  \\[1em]
  
  \begin{tabular}{@{}p{2em}p{0.2\linewidth}p{0.2\linewidth}p{0.2\linewidth}p{0.2\linewidth}@{}}
  \toprule
   \multirow{2}{*}{\makecell[lb]{Scene,\\ Round}} &\multirow{2}{*}{\makecell[lb]{Rule-Based\\ Nearest-Neighbor}} & \multicolumn{3}{c}{Scene2seq} \\
   \cmidrule{3-5}
  && Imitation learning & + Auxiliary Loss & + RL Fine-tuning \\
  \midrule
  B1 &\tellermsg{two finger spaces to right of helicopter , small hot air balloon . top of balloon level with helicopter propeller}
&\tellermsg{small hot air balloon , top right , 1 2 '' from top , 2 '' from right}&\tellermsg{small hot air balloon on right corner , fully visible .}
&\tellermsg{small hot air balloon top right}
\\
  \cmidrule{1-5}
B2 &\tellermsg{in front of tree is boy , he is to the left part of tree and is covering the curve up . he is angry , standing , arms , out facing left}
&\tellermsg{small girl , running , facing right , surprised , 1 '' from bottom , 1 2 '' from left}
&\tellermsg{small hot balloon on right corner , half '' from top .}
&\tellermsg{large bear on left faced right}
\\
\cmidrule{1-5}
B3&\tellermsg{the head of surprised girl is on front the trunk . she is like running and faces right .}
&\tellermsg{small pine tree behind her , bottom of trunk at horizon , bottom of trunk at horizon , small boy in front of tree , head touching bottom of tree , standing , smiling , facing right , holding a hot dog in left hand}
&\tellermsg{on center , a mad mike with hands front facing left .}
&\tellermsg{medium shocked girl running facing right holding hot dog in right hand}
\\
  \cmidrule{1-5}
  B4
  &\tellermsg{large bear on left side . facing right . arm cut off on left . feet half inch from bottom}
&\tellermsg{he is at the edge , small size , 1 2 '' from bottom , 1 2 '' from bottom , small kite in left corner , facing right}
&\tellermsg{on center , a small jenny , surprised , running facing right , shoulders at horizon .}
&\tellermsg{boy on right angry facing left holding kite in right hand hot dog in right hand}
\\
  \cmidrule{1-5}
B5
&\tellermsg{place a hot dog in her left hand}
& \texttt{<STOP>}
&\tellermsg{on center , a small hot balloon , half finger from right side .}
& \texttt{<STOP>}\\
  \cmidrule{1-5}
B6&\tellermsg{right top a kite , small size , tail is off screen . below kite dog medium size , facing left}&&\tellermsg{a small kite on right corner , only half visible .}&
\\
  \cmidrule{1-5}
B7&\texttt{<STOP>}&&\texttt{<STOP>}
\\
  \bottomrule
  \end{tabular}

  \caption{A comparison of the descriptions generated by each of our \Teller models for two randomly-sampled scenes from the test set.}
    \label{fig:appendix-qualitative-tellers}
\end{figure}

\end{document}